\documentclass[runningheads]{llncs}
\usepackage[T1]{fontenc}
\usepackage{amsmath}
\usepackage{amssymb}
\usepackage{caption}
\usepackage{subcaption}
\usepackage{graphicx}
\usepackage{tabularx}
\usepackage{booktabs}
\usepackage{float}
\usepackage{bbding}
\usepackage{hyperref}       
\usepackage{xcolor}
\usepackage{sidecap}
\hypersetup{
    colorlinks,
    linkcolor={blue!50!black},
    citecolor={blue!50!black},
    urlcolor={blue!50!black}
}

\begin{document}
\title{Evaluating the Explainability of Vision Transformers in Medical Imaging}
%
%

\author{Leili Barekatain \Envelope
\and
Ben Glocker
}

\authorrunning{L. Barekatain and B. Glocker}
%
\institute{
  Department of Computing, Imperial College London, UK\\
  \email{lb124@imperial.ac.uk}
}

\maketitle              
\begin{abstract}
Understanding model decisions is crucial in medical imaging, where interpretability directly impacts clinical trust and adoption. Vision Transformers (ViTs) have demonstrated state-of-the-art performance in diagnostic imaging; however, their complex attention mechanisms pose challenges to explainability. This study evaluates the explainability of different Vision Transformer architectures and pre-training strategies—ViT, DeiT, DINO, and Swin Transformer—using Gradient Attention Rollout and Grad-CAM. We conduct both quantitative and qualitative analyses on two medical imaging tasks: peripheral blood cell classification and breast ultrasound image classification. Our findings indicate that DINO combined with Grad-CAM offers the most faithful and localized explanations across datasets. Grad-CAM consistently produces class-discriminative and spatially precise heatmaps, while Gradient Attention Rollout yields more scattered activations. Even in misclassification cases, DINO with Grad-CAM highlights clinically relevant morphological features that appear to have misled the model. By improving model transparency, this research supports the reliable and explainable integration of ViTs into critical medical diagnostic workflows. The code is available at \url{https://github.com/leilibrk/Explainability-of-ViTs}.

\end{abstract}
\section{Introduction}
The integration of artificial intelligence into medical image analysis has led to significant advancements, particularly in biomedical image classification. Precise classification of medical images helps in early disease detection and reduces the risk of misdiagnosis. Although deep learning models excel in general image classification, biomedical images present challenges due to complex structures, varying noise levels, and intra- and inter-class variability. Convolutional Neural Networks (CNNs) have played a pivotal role in this domain by learning hierarchical features, yet struggle to capture long-range dependencies within images.

To overcome these limitations, Vision Transformers (ViTs) have emerged as a promising alternative, leveraging self-attention mechanisms to model global relationships across image patches. Unlike CNNs, ViTs process entire images holistically, leading to improved performance in complex classification tasks. However, a key challenge remains: the interpretability of these models.

Explainability is critical in medical applications, where clinicians must validate model predictions to ensure they are based on meaningful biological features rather than spurious correlations. Relying solely on performance metrics such as accuracy is insufficient; we must also apply explainability techniques to understand and trust model decisions. Current interpretability techniques, such as attention-based and feature attribution methods, provide insights into model decision-making but vary in effectiveness across different architectures. 

This study aims to bridge this gap by evaluating different Vision Transformer architectures and pre-training strategies, namely ViT, DeiT, DINO, and Swin Transformer on  peripheral blood cell classification and breast ultrasound image classification tasks. We assess their explainability using Gradient Attention Rollout and Grad-CAM, analyzing their ability to provide reliable, interpretable predictions. We investigate (1) how effectively these models classify images and
(2) how well different explainability methods reveal their decision-making process. Our findings contribute to the growing field of AI-driven medical diagnostics by shedding light on the balance between model accuracy and interpretability.

\subsection{Related Work}
Recent works have provided valuable insights into making Vision Transformers interpretable in the context of medical imaging. Komorowski et al. \cite{komorowski2023towards} introduced metrics—faithfulness, sensitivity, and complexity—to benchmark model explainability. They conducted experiments on chest X-ray classification, comparing methods like LIME~\cite{ribeiro2016should}, Attention Rollout~\cite{abnar2020quantifying}, and TransLRP~\cite{chefer2021transformer}. However, their analysis was limited to fine-tuning a base-sized ViT model and did not explore how different ViT architectures may affect explainability.

Several studies have also developed explainable ViTs for specific medical tasks. xViTCOS \cite{mondal2021xvitcos} is one such model, designed for COVID-19 screening using chest X-rays. It employs Gradient Attention Rollout~\cite{gil2021vitexplain} to assess model explainability. The results show that the model not only outperforms recent benchmarks but also attends to meaningful image regions, as confirmed by radiologists. However, the study assesses explainability qualitatively through visual alignment with clinical expectations and does not provide quantitative metrics to objectively compare the attribution quality of the generated saliency maps. A related study~\cite{hroub2024explainable} also focused on COVID-19 and pneumonia prediction from chest X-rays using different explainable deep learning models, including ViTs. The system integrates Grad-CAM~\cite{selvaraju2017grad} to visualize critical regions influencing predictions and reports high diagnostic accuracy across multiple lung disease categories. However, it also lacked a quantitative evaluation of explainability.

\section{Methods}
 
The \textbf{Vision Transformer (ViT)}~\cite{alexey2020image} divides an image into fixed-size non-overlapping patches, embeds them linearly, and processes them using a transformer encoder to capture global context. Each encoder block uses self-attention, which allows every patch to attend to all others. This enables the model to learn both local and global dependencies. ViTs have shown strong performance in medical imaging tasks such as disease classification and lesion detection, often outperforming CNNs. However, standard ViTs often require large labeled datasets, which can limit their use in medical applications.

To reduce the reliance of ViTs on large-scale datasets, the \textbf{Data-efficient Image Transformer (DeiT)} was proposed  \cite{touvron2021training}. DeiT uses a student-teacher distillation approach, where a CNN teacher guides training via a distillation token that interacts with class and patch tokens. This helps the model learn from the teacher’s predictions and achieve strong performance on smaller datasets. Sevinc et al.~\cite{sevinc2025distillation} utilized DeiT architecture for classifying brain MRI images using a small dataset of 3,264. The model achieved a notable accuracy of 93.69\%.

The \textbf{DINO (Self-Distillation with No Labels)} framework \cite{caron2021emerging} is a self-supervised approach that trains ViT without labels using a teacher–student model, where both networks share the same architecture and process different augmented views. The teacher is updated using an exponential moving average of the student, allowing the model to learn meaningful visual representations without explicit supervision. Cisternino et al.~\cite{cisternino2024self} used DINO to classify histopathological images across 23 tissue types. The model extracted meaningful features without manual labels and outperformed other self-supervised methods.

\textbf{Swin Transformer} \cite{liu2021swin} improves ViTs by introducing hierarchical feature representation and shifted window attention, reducing computational complexity from quadratic to linear. Self-attention is applied within local windows, and shifting these windows across layers enables global context modeling. Jin et al.~\cite{jin2025multitask} proposed the Multitask Swin Transformer (MTST) for pulmonary nodule analysis in CT images. This multitask approach, which predicts malignancy and characterization scores concurrently, provides clinicians information similar to a radiologist’s assessment and improves the model’s practical diagnostic utility.

\subsection{Explainability Techniques}
As Vision Transformers rapidly advance medical imaging, understanding their decision-making has become crucial. Here, we discuss two common methods for interpreting ViT decisions. 

\subsubsection{Attention-Based Methods}

Attention-based methods aim to explain Vision Transformers by analyzing how attention is distributed across image tokens. Since attention mechanisms allow the model to assign different importance levels to different regions of the input, they offer a natural way to explore interpretability~\cite{vaswani2017attention}. One widely used technique is Attention Rollout \cite{abnar2020quantifying}, which tracks how information flows through the transformer layers by recursively multiplying averaged attention matrices:


\begin{equation}\label{eq:rollout2}
    rollout = {\hat{A}}^{(1)} \cdot~ {\hat{A}}^{(2)} \cdot  ... \cdot {\hat{A}}^{(B)}
\end{equation}

However, Attention Rollout produces the same explanation regardless of the predicted class, which limits its ability to highlight class-specific features \cite{kashefi2023explainability}. 
To address this, \textbf{Gradient Attention Rollout} \cite{gil2021vitexplain} weights attention layers using gradients to highlight regions of the image that contributed to the model’s decision. To ensure class-specificity, attention values are multiplied by corresponding gradients. The final attention map is obtained by averaging across attention heads, preserving only most relevant and positively contributing attention paths for the target class. In this way, Gradient Attention Rollout produces class-specific attention maps that better reflect the model’s decision-making process.

\subsubsection{Feature Attribution Methods}
Unlike attention-based techniques that analyze the internal model mechanisms, feature attribution methods explain model predictions by identifying which input features most influenced the output. \textbf{Grad-CAM} \cite{selvaraju2017grad} is a gradient-based feature attribution method that computes the gradient of the target class score with respect to the activations of a selected transformer layer, averaging these gradients to get importance weights for each feature map, and combining them to generate a heatmap. ReLU activation keeps only positively contributing features, and the result is upsampled to the input resolution. The final output is a class-discriminative heatmap that highlights regions most responsible for the model’s decision. In medical imaging, such visual explanations are essential to ensure models rely on clinically meaningful features.

\section{Experiments \& Results}
\subsection{Datasets and Implementation Details}

\textbf{Peripheral Blood Cell (PBC) Dataset} ~\cite{acevedo2020dataset} is a public dataset that contains 17,092 high-resolution (360×363) JPEG images of eight categories of peripheral blood cells, including Basophil, Eosinophil, Erythroblast, Immature Granulocyte, Lymphocyte, Monocyte, Neutrophil, and Platelet. 

\textbf{Breast Ultrasound Images Dataset}~\cite{al2020dataset} is a public dataset containing 780 PNG images (500×500). The data were collected in 2018 from 600 female patients, aged between 25 and 75 years. The images are categorized into three classes: normal, benign, and malignant.

\textbf{Implementation Details} All images were resized to 224×224, center-cropped, and normalized, with no additional preprocessing or augmentation.

We fine-tuned four pre-trained Vision Transformer models from the \texttt{PyTorch Transformers} library: ViT (google/vit-base-patch16-224), DeiT (facebook/deit-tiny-patch16-224), DINO-ViT (facebook/dino-vits16), and Swin (microsoft/swin-tiny-patch4-window7-224). Training was performed with a batch size of 32. Given that these models were extensively pre-trained on large-scale datasets, and considering the relatively modest size of the PBC dataset, one epoch was sufficient to fine-tune the models effectively without risking overfitting. However, for the breast ultrasound image classification dataset, due to its smaller size (780 images), we experimented with different settings and found that training the models for 10 epochs was sufficient to achieve stable performance without overfitting. Throughout the training and evaluation phases, we monitored performance metrics, including accuracy and F1-score. After training, Gradient Attention Rollout and Grad-CAM were applied to analyze the explainability of model predictions.

\subsection{Performance Results}
The performance of the four vision transformer architectures, including ViT, DeiT, DINO-ViT, and Swin was evaluated using standard classification metrics for two datasets. Table \ref{tab:performance_metrics_blood} and \ref{tab:performance_metrics_breast} present the results.
\begin{table}[ht]
\centering
\begin{minipage}[t]{0.48\textwidth}
    \centering
    \begin{tabular}{|l|c|c|}
        \hline
        Model & Accuracy (\%) & F1-score (\%) \\
        \hline
        ViT & 98.68 & 98.73 \\
        \hline
        DeiT & 98.05 & 97.92 \\
        \hline
        DINO & 96.97 & 97.16 \\
        \hline
        Swin & 98.58 & 98.59 \\
        \hline
    \end{tabular}
    \caption{Results on PBC}
    \label{tab:performance_metrics_blood}
\end{minipage}%
\hfill
\begin{minipage}[t]{0.48\textwidth}
    \centering
    \begin{tabular}{|l|c|c|}
        \hline
        Model & Accuracy (\%) & F1-score (\%) \\
        \hline
        ViT & 87.18 & 85.66 \\
        \hline
        DeiT & 79.49 & 75.48  \\
        \hline
        DINO & 80.77 & 77.23 \\
        \hline
        Swin & 89.74 & 88.44 \\
        \hline
    \end{tabular}
    \caption{Results on Breast Ultrasound}
    \label{tab:performance_metrics_breast}
\end{minipage}
\end{table}

\subsection{Explainability Results—Quantitative}

Insertion and Deletion are complementary metrics for evaluating the faithfulness of an explanation method. In the Insertion metric, the most important pixels from the generated heatmaps are gradually added to a blank or blurred image, and the probabilities of the target class are tracked as pixels are inserted. A steeper increase in confidence indicates that the explanation has accurately identified the most relevant regions. Conversely, in the Deletion metric, the most important pixels are progressively masked from the original image, and the corresponding drop in target class probabilities is recorded. A more rapid decline suggests higher explanation fidelity~\cite{kashefi2023explainability,petsiuk2018rise}. 

\begin{figure}
    \centering
    \begin{subfigure}[t]{0.48\linewidth}
        \includegraphics[width=\linewidth]{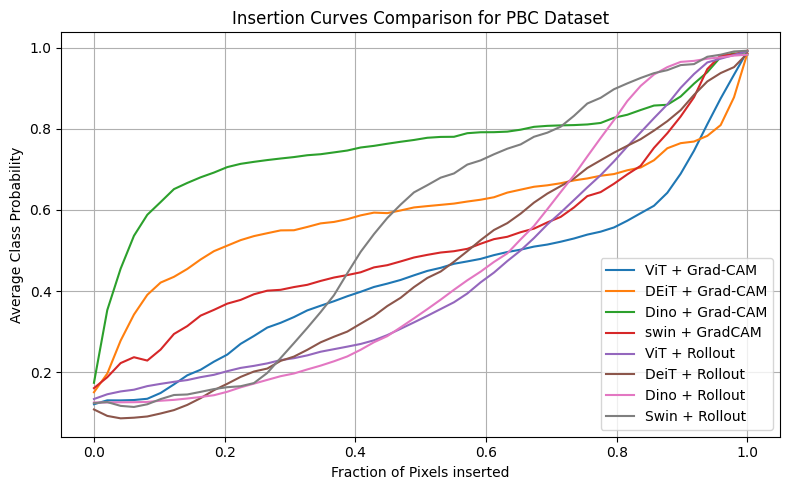}
        \caption{}
        \label{fig:insertion-blood}
    \end{subfigure}
    \hfill
    \begin{subfigure}[t]{0.48\linewidth}
        \includegraphics[width=\linewidth]{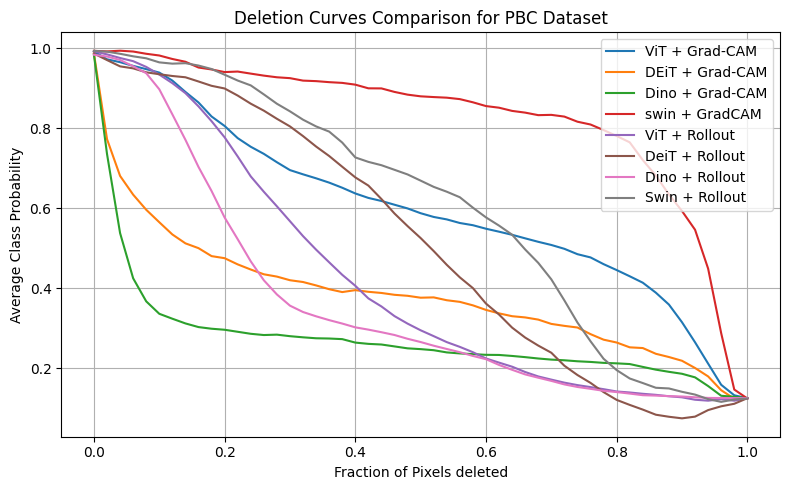}
        \caption{}
        \label{fig:deletion-blood}
    \end{subfigure}
    \begin{subfigure}[t]{0.48\linewidth}
        \includegraphics[width=\linewidth]{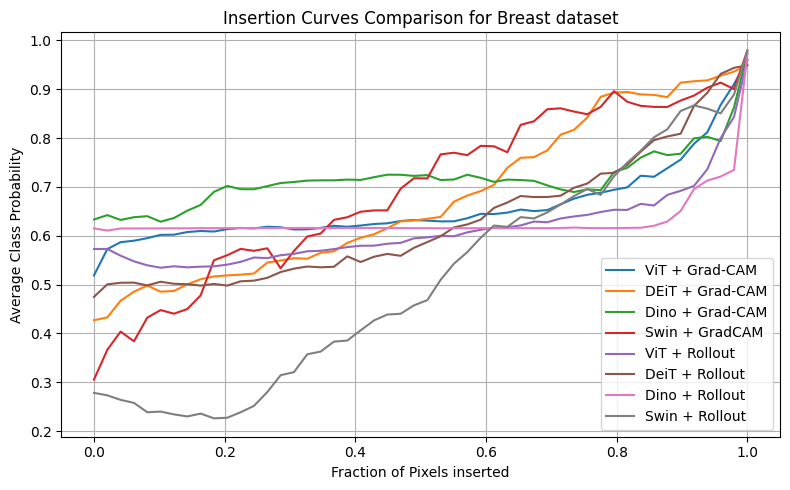}
        \caption{}
        \label{fig:insertion-breast}
    \end{subfigure}
    \hfill
    \begin{subfigure}[t]{0.48\linewidth}
        \includegraphics[width=\linewidth]{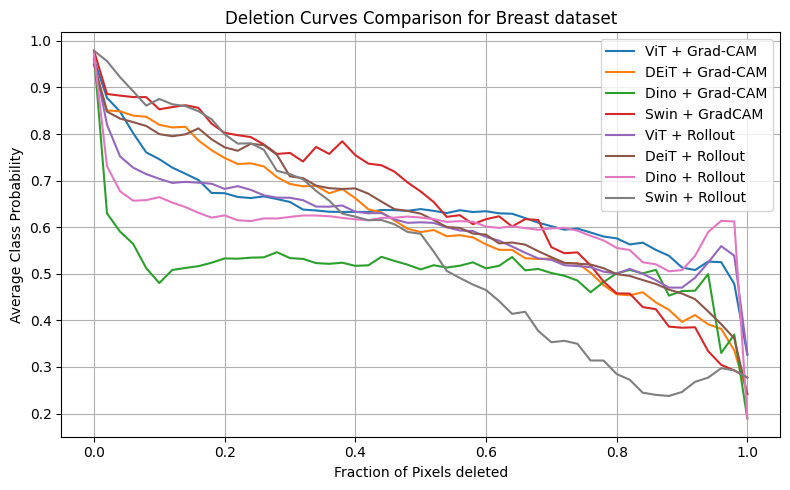}
        \caption{}
        \label{fig:deletion-breast}
    \end{subfigure}
    \caption{Visualization of (a,c) inserting and (b,d) deleting the most relevant pixels—identified by Grad-CAM and Gradient Attention Rollout—on the predicted class probability across four Vision Transformer models for the PBC dataset (a,b) and Breast Ultrasound dataset (c,d).}
    \label{fig:insertion-deletion}
\end{figure}

We evaluated the explainability of Gradient Attention Rollout and Grad-CAM across four architectures using these metrics. For the PBC dataset, we randomly sampled 520 validation images (65 per class). For the Breast Ultrasound dataset, we used all 117 validation images. Both methods were applied uniformly across models and Insertion and Deletion AUC were used to measure how well explanations matched the models’ decisions. Results in Figure~\ref{fig:insertion-deletion} show that Grad-CAM consistently outperforms Gradient Attention Rollout across all architectures, showing higher AUC in insertion (faster confidence recovery) and lower AUC in deletion (steeper confidence drop), indicating better localization of critical visual features.

\begin{table}
    \centering
    \small
    \begin{tabular}{lc|c|c|c|c|c|c|c|}
        \toprule
        & \multicolumn{4}{c}{\textbf{Grad-CAM}} & \multicolumn{4}{c}{\textbf{Rollout}} \\
        \cmidrule(lr){2-5} \cmidrule(lr){6-9}
        & ViT & DeiT & DINO & Swin & ViT & DeiT & DINO & Swin \\
        \midrule
        Deletion(↓)  &   0.60  &  0.38   &  \textbf{0.27}   &  0.82   &  0.42   &   0.52  &  0.36   & 0.60    \\
        Insertion(↑) &  0.44   &  0.60   &   \textbf{0.75}  &   0.52  &  0.44   &  0.45   &   0.45  & 0.56    \\
        \bottomrule
    \end{tabular}
    \caption{The Deletion and Insertion AUC across models using Grad-CAM and Rollout explanations for PBC dataset.}
    \label{tab:blood-metric}
\end{table}
\begin{table}
    \centering
    \small
    \begin{tabular}{lc|c|c|c|c|c|c|c|}
        \toprule
        & \multicolumn{4}{c}{\textbf{Grad-CAM}} & \multicolumn{4}{c}{\textbf{Rollout}} \\
        \cmidrule(lr){2-5} \cmidrule(lr){6-9}
        & ViT & DeiT & DINO & Swin & ViT & DeiT & DINO & Swin \\
        \midrule
        Deletion(↓)  &  0.63   &  0.61   &  \textbf{0.51}   & 0.65   &  0.61   &   0.63  & 0.60    & 0.56    \\
        Insertion(↑) &  0.65   & 0.67   & \textbf{0.72}  & 0.69  & 0.61 &  0.62   &  0.62   &  0.50 \\
        \bottomrule
    \end{tabular}
    \caption{The Deletion and Insertion AUC across models using Grad-CAM and Rollout explanations for Breast Ultrasound dataset.}
    \label{tab:Breast-metric}
\end{table}
The Insertion and Deletion AUCs are presented in Table~\ref{tab:blood-metric} and Table~\ref{tab:Breast-metric}. Based on the results, DINO gives the best AUC scores for both datasets, suggesting that it is easier to explain this model's decisions using Grad-CAM.
\subsection{Explainability Results - Qualitative}
\begin{figure}[ht]
  \centering
  \begin{minipage}[t]{0.48\linewidth}
    \centering
    \includegraphics[width=\linewidth]{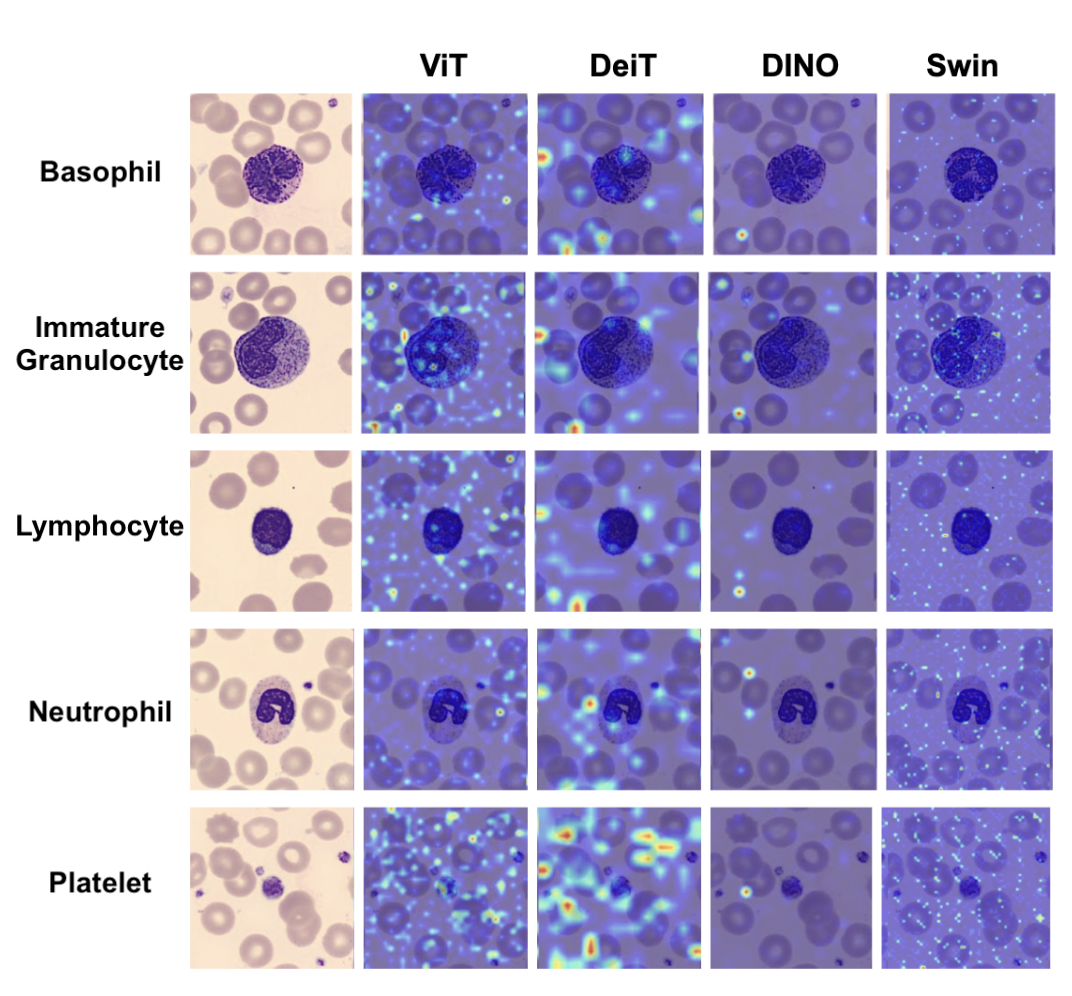}
    \caption*{(a) Gradient Attention Rollout}
  \end{minipage}
  \hfill
  \begin{minipage}[t]{0.48\linewidth}
    \centering
    \includegraphics[width=\linewidth]{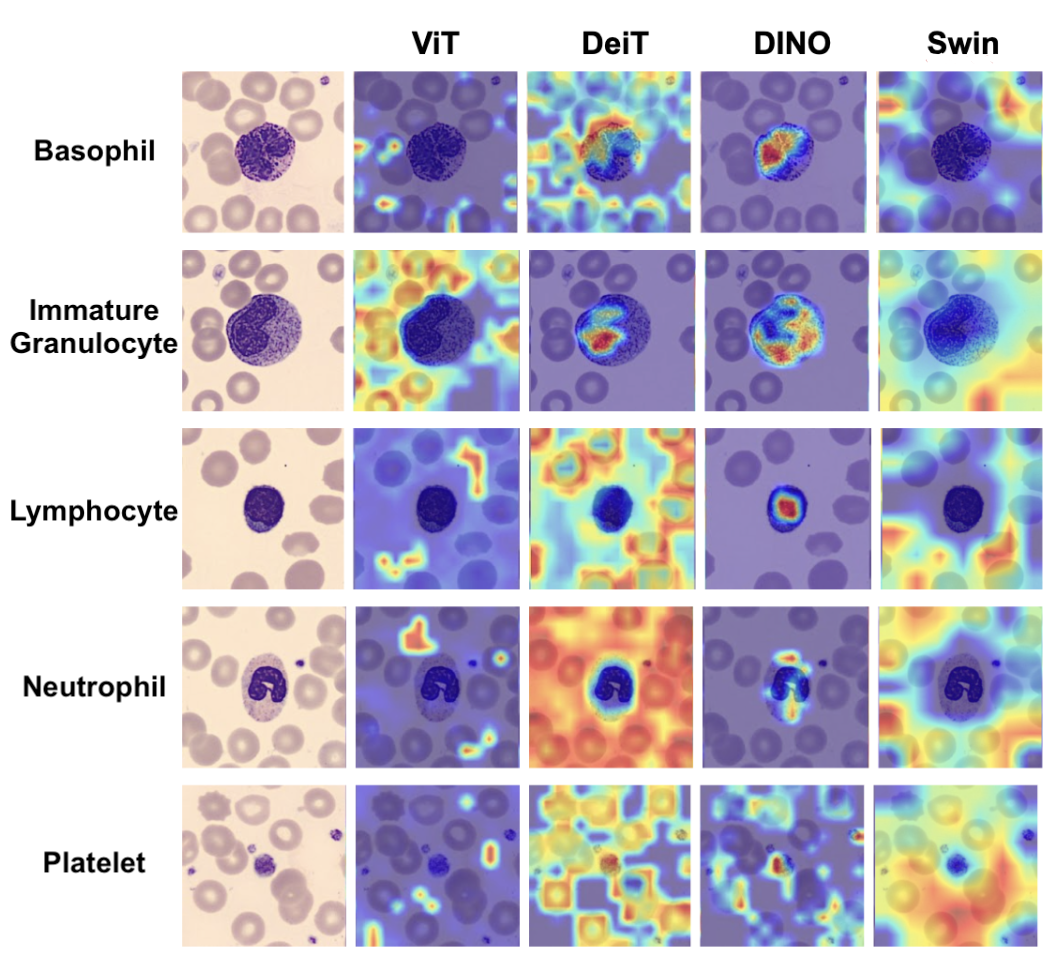}
    \caption*{(b) Grad-CAM}
  \end{minipage}
  \caption{Comparison of Gradient Attention Rollout and Grad-CAM heatmaps for five blood cell classes from the PBC dataset across four models. First column shows the input images, followed by heatmaps of the predicted class. }
  \label{fig:qualitative-comparison-blood}
\end{figure}

\begin{figure}[ht]
  \centering
  \begin{minipage}[t]{0.495\linewidth}
    \centering
    \includegraphics[width=\linewidth]{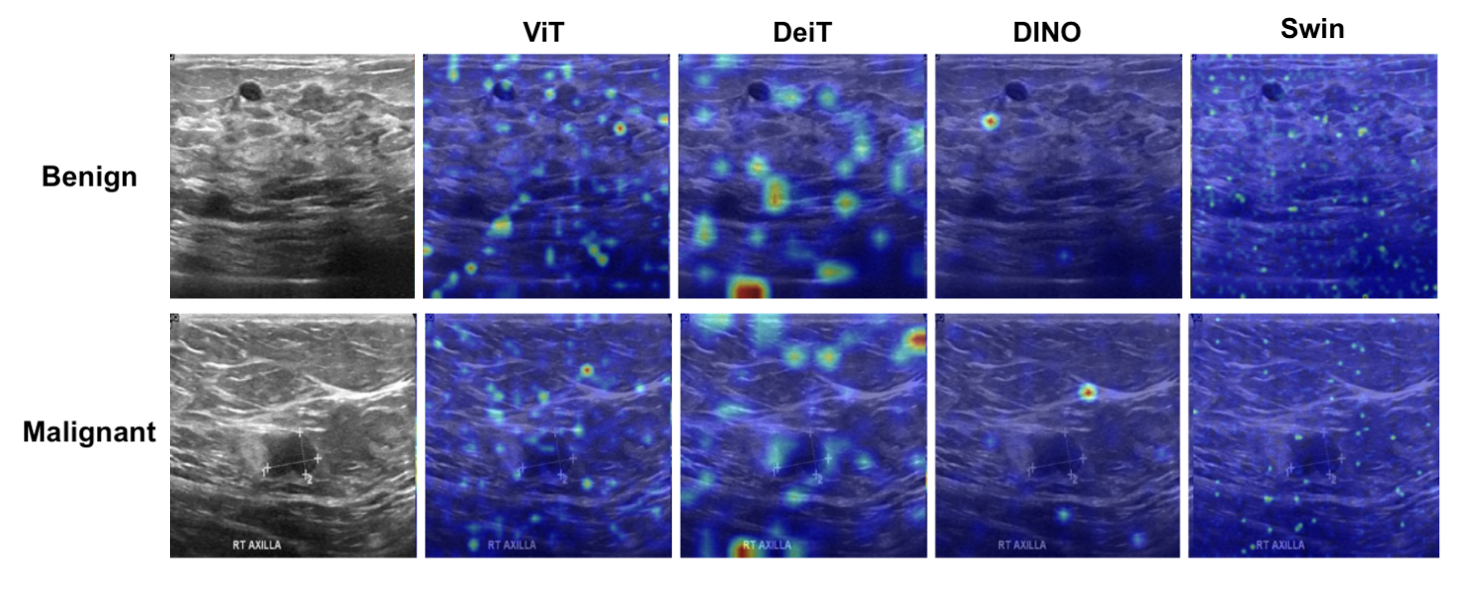}
    \caption*{(a) Gradient Attention Rollout}
  \end{minipage}
  \begin{minipage}[t]{0.495\linewidth}
    \centering
    \includegraphics[width=\linewidth]{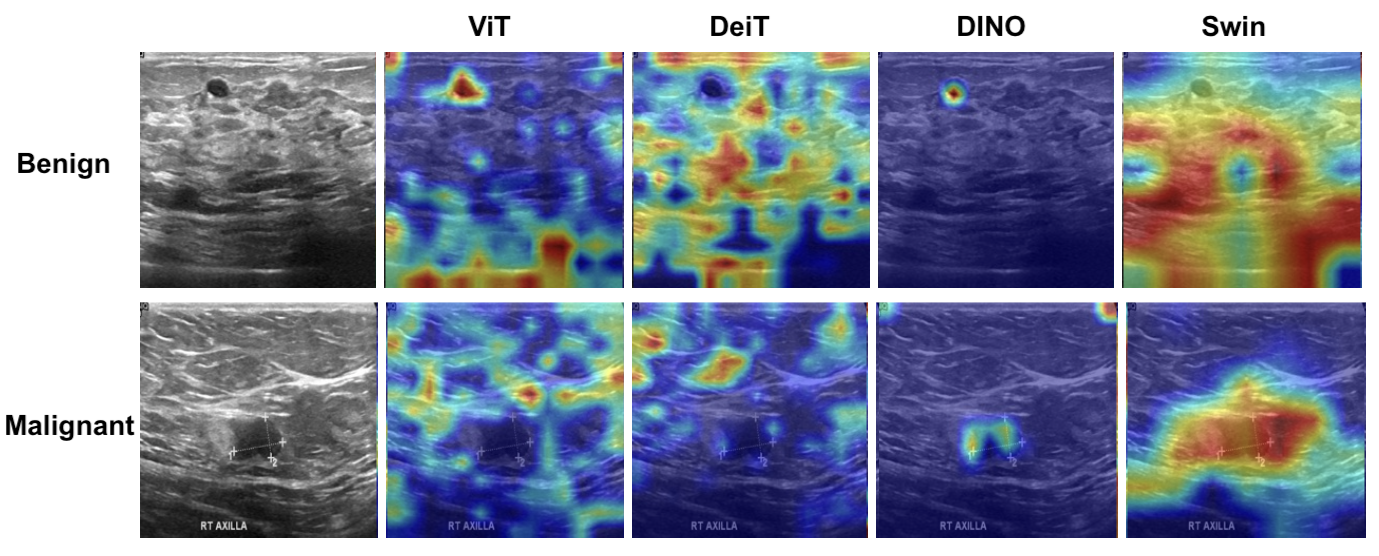}
    \caption*{(b) Grad-CAM}
  \end{minipage}
  \caption{Comparison of Gradient Attention Rollout and Grad-CAM heatmaps for benign and malignant breast ultrasound images across four models. First column shows the input images, followed by heatmaps of the predicted class.}
  \label{fig:qualitative-comparison-breast}
\end{figure}

Figures~\ref{fig:qualitative-comparison-blood} and~\ref{fig:qualitative-comparison-breast} show the qualitative results of Gradient Attention Rollout and Grad-CAM across four transformer models on the PBC and breast ultrasound datasets. In general, Grad-CAM produces more focused and interpretable heatmaps, with stronger activation over the relevant regions. However, Gradient Attention Rollout produces scattered and inconsistent attention maps that often highlight irrelevant background regions, making its explanations more difficult to interpret. Among the models, DINO-ViT combined with Grad-CAM consistently provides the most coherent and clinically meaningful attributions.

For example, in Figure~\ref{fig:qualitative-comparison-blood} \textit{right} in the Basophil case (first row), DINO with Grad-CAM highlights the entire cell body accurately, while other models show more dispersed or weaker activations. Similarly, in the Immature Granulocyte case (second row), DINO with Grad-CAM shows a precise heatmap centered over the nucleus and cytoplasmic granules. 

In Figure~\ref{fig:qualitative-comparison-breast} \textit{right}, 
it is evident that DINO with Grad-CAM again demonstrates superior explainability. In the benign case (first row), it produces a focused activation map centered on the lesion boundary, while other models such as DeiT and ViT display scattered attention across the entire image, including irrelevant tissue regions. Swin shows stronger activation but extends beyond the lesion area. In the malignant case (second row), DINO localizes the irregularly shaped mass, focusing precisely on the contours of the tumor. 

\subsubsection{Qualitative Error Analysis}
\begin{figure}[t]
  \centering
  \includegraphics[width=\linewidth]{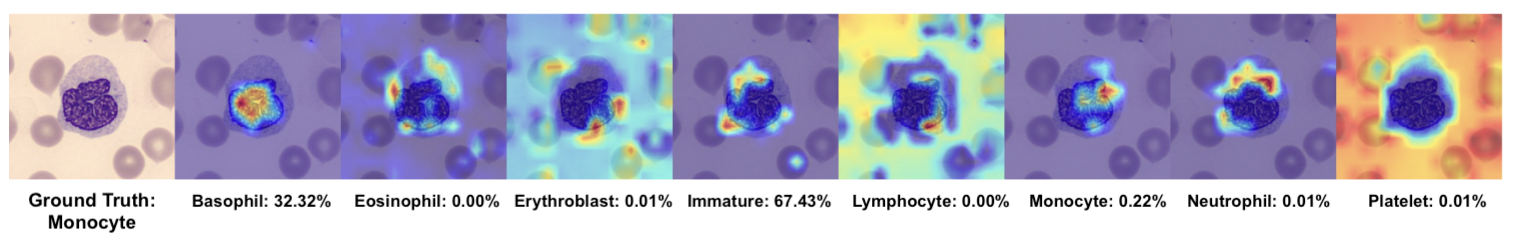}
  \includegraphics[width=0.5\linewidth]{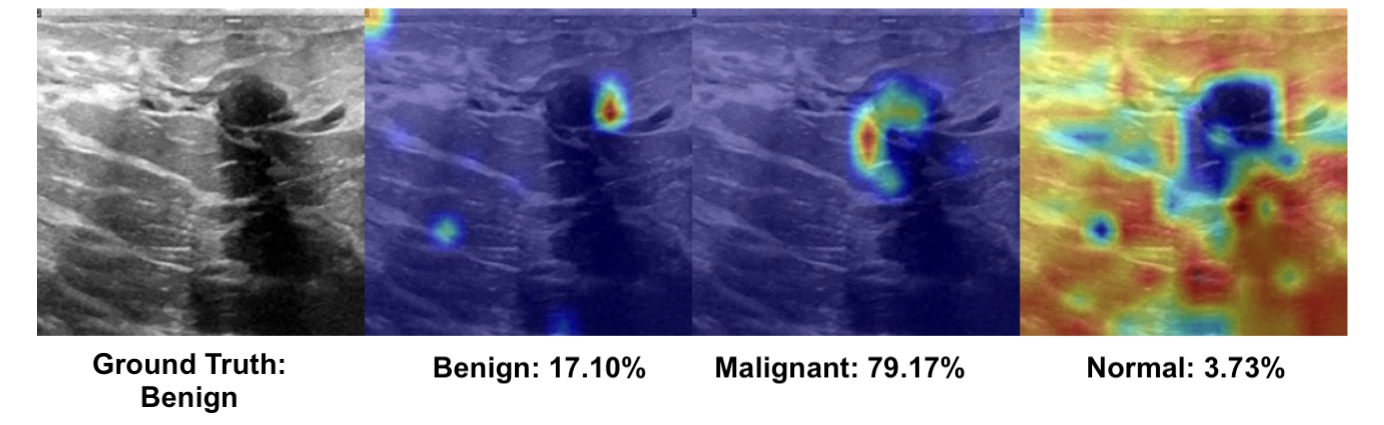}
  \caption{Grad-CAM visualizations of misclassified images by the DINO-ViT model. \textbf{Top:} PBC dataset — ground truth is Monocyte, but predicted as Immature Granulocyte with 67.43\% confidence. \textbf{Bottom:} Breast ultrasound — ground truth is Benign, but predicted as Malignant with 79.17\% confidence.}
  \label{fig:misclassification-dino}
\end{figure}

When a model is explainable, its visual attributions—such as Grad-CAM heatmaps—can provide insights into the reasoning behind incorrect predictions. Figure~\ref{fig:misclassification-dino} shows two such cases by the DINO-ViT model. In the PBC example (top), the model predicted the class as \textit{Immature Granulocyte} with 67.43\% confidence, while the ground truth label is \textit{Monocyte}. The Grad-CAM heatmap for the predicted class shows strong activation around the nucleus, suggesting the model misinterpreted the shape and granularity as irregular features typical of Immature Granulocyte cells, which led to the misclassification. Meanwhile, the heatmap for the true class displays weaker and more diffused attention, suggesting that the model has not learned a sufficiently strong representation for Monocyte. The activation patterns observed across other classes further demonstrate how subtle visual similarities can lead to confusion in fine-grained classification tasks. 

In the breast ultrasound example (bottom), the model misclassified a \textit{Benign} lesion as \textit{Malignant} with 79.17\% confidence. The heatmap for the predicted class highlights the central lesion area with strong intensity, indicating that the model focused on structural features associated with malignancy, such as asymmetry or irregular margins. Meanwhile, the heatmap for the true class appears weaker and less defined, suggesting the model failed to learn sufficiently discriminative features for benign cases. These examples demonstrate that explainability methods like Grad-CAM can reveal the underlying reasons behind model misclassifications, offering valuable insights for diagnosing and improving model behavior.

\section{Conclusion and Future Works}
This study aimed to provide a systematic framework for identifying the most explainable Vision Transformer architecture and explanation method for medical imaging tasks. We evaluated four architectures—ViT, DeiT, DINO-ViT, and Swin Transformer—using two explainability techniques: Grad-CAM and Gradient Attention Rollout. Our experiments on peripheral blood cell (PBC) and breast ultrasound datasets revealed that while all models achieved high classification accuracy, their interpretability varied notably. Grad-CAM consistently provided more localized and class-discriminative explanations compared to Gradient Attention Rollout. When using Grad-CAM as the benchmark for explainability, DINO emerged as the most interpretable setup. In contrast, ViT, DeiT, and Swin Transformer showed scattered and less focused attention, which may limit their clinical applicability. Even in misclassification cases, DINO’s attention maps still highlighted relevant morphological features, helping to uncover potential reasons behind the model’s decision-making. Interestingly, ViT and Swin outperformed DINO-ViT in terms of accuracy and F1-score across both datasets (see Tables~\ref{tab:performance_metrics_blood} and \ref{tab:performance_metrics_breast}). However, DINO consistently provided more explainable results. This suggests that model selection in critical domains like medical diagnostics should not rely solely on performance metrics, but also consider the quality of explanations. 

Future work could focus on developing more accurate and clinically meaningful explainability methods specifically designed for ViTs. Designing hybrid techniques that combine spatial precision with deeper semantic understanding could enhance interpretability. Additionally, incorporating domain-specific priors or medical constraints into the explanation process may improve faithfulness.  

\begin{credits}

\subsubsection{\discintname}
The authors have no competing interests to declare.
\end{credits}

%
%
\bibliographystyle{splncs04}
\bibliography{reference}

\end{document}